\documentclass[10pt, conference, compsocconf]{IEEEtran}
\ifCLASSINFOpdf
  \usepackage[pdftex]{graphicx}
\else
\fi
\begin{document}
%
\title{Second order scattering descriptors predict fMRI activity due to visual textures}


\author{\IEEEauthorblockN{Michael Eickenberg \\ Fabian Pedregosa}
\IEEEauthorblockA{Inria Parietal\\
Neurospin, CEA Saclay\\
France\\
michael.eickenberg@nsup.org \\ fabian.pedregosa@inria.fr}

\and
\IEEEauthorblockN{Mehdi Senoussi}
\IEEEauthorblockA{CNRS Toulouse\\
Toulouse, France\\
Email: name@xyz.com}
\and
\IEEEauthorblockN{Alexandre Gramfort}
\IEEEauthorblockA{T\'el\'ecom Paris-Tech, CNRS\\
Paris, France\\
Email: name@xyz.com}
\and
\IEEEauthorblockN{Bertrand Thirion}
\IEEEauthorblockA{Inria Parietal\\
Neurospin, CEA Saclay\\
France\\
Email: bertrand.thirion@inria.fr}
}


%


\maketitle

\begin{abstract}
Second layer scattering descriptors are known to provide good classification performance on natural quasi-stationary processes such as visual textures due to their sensitivity to higher order moments and continuity with respect to small deformations. 
In a functional Magnetic Resonance Imaging (fMRI) experiment we
present visual textures to subjects and evaluate the predictive power
of these descriptors with respect to the predictive
power of simple contour energy - the first scattering layer.
%
We are able to conclude not only that invariant second layer scattering coefficients better encode voxel activity, but also that well predicted voxels need not necessarily lie in known retinotopic regions.
\end{abstract}

\begin{IEEEkeywords}
fMRI; scattering; encoding; retinotopy

\end{IEEEkeywords}

%
\IEEEpeerreviewmaketitle

\section{Introduction}

Advances in vision research using fMRI have shown that the Blood Oxygen-Level Dependent (BOLD) fMRI response to natural images is well explained as a general linear model of the energies of a Gabor filter pyramid at well chosen scales, orientations and locations. 
Apart from reflecting general retinotopy due to the local nature of the filter shape, it has been shown that the energy in local edges contributes significantly to the signal. 
This finding is expected in the sense that, at least in the earliest cortical visual area, V1, a large number of neurons is known to actually perform the task of edge detection, be it in a phase dependent (simple cells) or independent (complex cells) manner. The discovery of this kind of tuning in primary visual neurons dates back to Hubel and Wiesel \cite{HubelWiesel1968}.
In effect, the brain regions, of which the BOLD signal is best modeled using a Gabor energy model, can be found in V1.

Moving away from V1, this model becomes less and less plausible as an explanation of neural activity. 
While edge detectors still may be present in higher level visual areas, other, more complex functionalities are added, and one might wish to model these more accurately. 
In an attempt to introduce more complex features, the authors of \cite{EickenbergPRNI2012} modelled the voxel responses using a windowed scattering transform, which, in addition to a first layer of Gabor moduli and smoothing, provides a second layer of wavelet transforms on the initial moduli, followed again by the nonlinear complex modulus and smoothing. 
%
%
This result was further examined in a subsequent work \cite{EickenbergICMLSTAMLINS2012} by the finding that filter choice plays an important role.
%

%

\begin{figure}[h]
  \begin{centering}
    \includegraphics[width=0.9\linewidth]{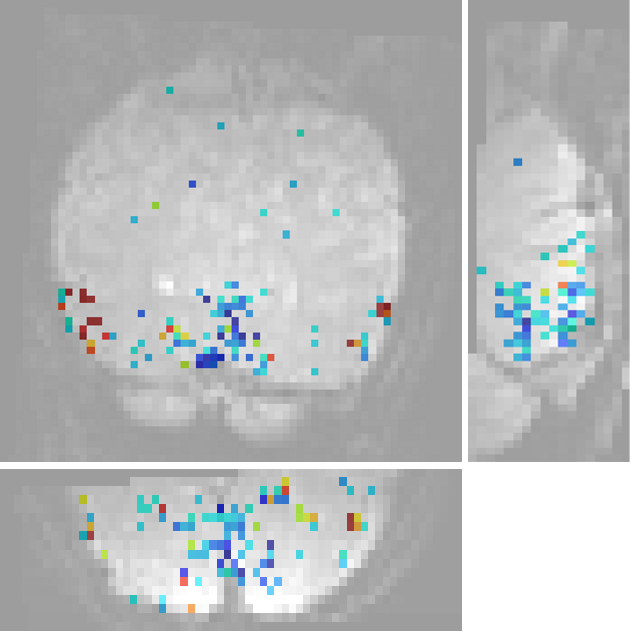}
  \end{centering}
  \caption{Difference of predictive \(r^2\) scores between two models shown on mean fMRI image of the dataset in \cite{Kay2008}: Blue means better fit using a simple Gabor pyramid: scattering level 1. Red shows regions where scattering level 2 adds predictive power. These regions are outside the primary visual areas. Top: oblique coronal; Middle: oblique axial; Bottom: sagittal}\label{fig:gall}
\end{figure}

Zero-sum Morlet wavelets improve the fit of first order descriptors to an extent that they compete with second order features in some regions, but are outperformed in others:
As can be seen in Fig. \ref{fig:gall}, there exist spatially contiguous regions largely outside commonly mapped retinotopic areas, in which adding second layer scattering coefficients is beneficial to the model.

Simple but effective object recognition mechanisms employ histograms of oriented gradients (HOG \cite{Dalal2005}) at several scales followed by a classifier, such as SVM. The fact that this type of setup works shows the importance of contours in the processing of general natural images. HOG is similar to DAISY \cite{daisy2007} which corresponds to the first layer of the scattering transform.

However, the performance of this simple contour detection setup in the classification of textures is poorer. Natural textures can be seen as (mostly non-Gaussian) stationary processes in which features capturing higher order moments have been shown to be important for establishing a generative model: In \cite{Portilla2000}, multiplicative interactions between first order descriptors are used among other features. In \cite{Bruna2011}, \cite{Bruna2012} and \cite{Sifre2012}, second level scattering coefficients capture similar interactions. It turns out that certain 4th order moments carry important information about the visual aspect of natural textures and the mentioned higher order features correlate with some of them.
Indeed, a rotation invariant version of the scattering transform \cite{Sifre2012} linearises stationary textures to a degree that a K-flats type classifier \cite{KFlats}, a Gaussian mixture model or linear one-vs-rest classifiers perform well in distinguishing between classes, indicating that the semantic information texture class" has been made linearly separable.

The difficulty of capturing the statistics of visual textures in image processing led us to the question of how the brain may accomplish this task. We devised an experimental paradigm, where BOLD fMRI activity is recorded while subjects view images of natural textures and propose to analyse it using an encoding model \cite{Naselaris2011} based on the translation invariant version of the scattering transform.

\section{Methods}
The fMRI BOLD response to visual stimuli was acquired during a visual comparison task, where subjects were asked to distinguish between images of six different texture classes.

\subsection{The experimental task}

\begin{figure}
  \begin{centering}
    \includegraphics[width=.99\linewidth]{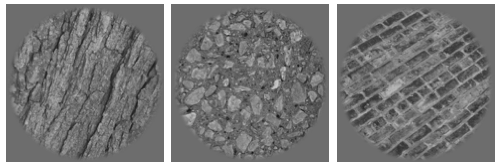}
    \includegraphics[width=.99\linewidth]{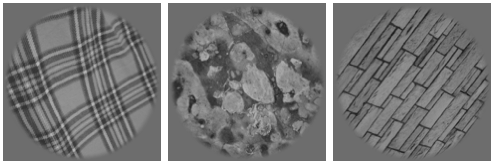}
  \end{centering}
  \caption{Sample stimuli used in the experiment: Extracts from the UIUC dataset \cite{Lazebnik05asparse}}
 \end{figure}

Subjects were asked to compare two texture instances from the same class, presented one after the other, while fixating a central cross. One experimental block took 12 seconds: At second 0, the first image was presented for one second, flashed three times in an on-off-on-off-on sequence of 200ms duration each. At second 4 the second image was presented in the same manner. At second 8, a smaller image, centered around the fixation point was presented, containing an extract of either the first, the second, or an unrelated image. The subject was asked to press a left-hand button if the first image had been repeated, the right-hand button if the second image had been repeated and no button if the image was unrelated.

\begin{figure}
  \begin{centering}
    \includegraphics[width=.9\linewidth]{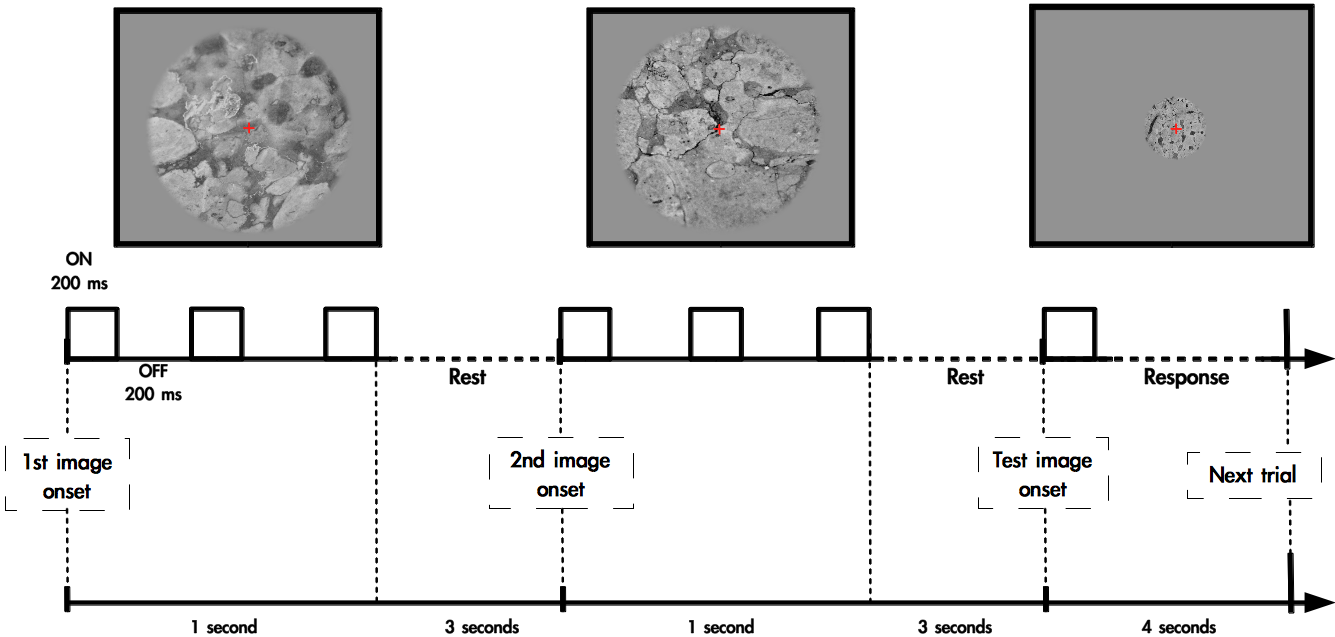}
  \end{centering}
  \caption{Visualisation of one 12 second block presenting 2 full images and one task related image extract.}
\end{figure}

One experimental session consisted of 36 such 12s blocks, which corresponds to the presentation of 72 images. All texture images were presented twice: once in first position, once in second position, in order to be able to account for effects due to ordering and in order to increase the signal to noise ratio in subsequent analysis.

One scanner session consisted of 6 experimental sessions. Thus a total of 216 distinct texture images were shown.

The task was presented to the subjects once before entering the scanner, with stimuli not used during the acquisition.

\subsection{Measurements}
Functional images were acquired on a 3T Siemens scanner (TR=2400ms, TE=30ms, matrix size \(128\times 128\), FOV 192mm\(\times\)192mm). Each volume consisted of 34 2mm-thick axial slices without gap. Anatomical T1 images were acquired on the same scanner with a spatial resolution of \(1\textrm{mm}\times 1\textrm{mm}\times 1\textrm{mm}\). 
EPI data acquisition was performed using the sequence in \cite{Boegle2010} with IPAT=2; this sequence includes distortion and motion correction at the acquisition level.
Slice timing correction and coregistration to the anatomy were performed using SPM8 software. 

Data were acquired from three subjects in two sessions each, using two different stimulus sets for the two sessions. Additionally, a fast retinotopic mapping procedure was carried out to obtain brain regions that display strong tuning to retinotopy. 

\subsection{Data preprocessing}

In order to slightly reduce the dimensionality and raise the signal to noise ratio, preprocessing of the time courses was performed in the form of a general linear model (GLM). This yielded an activity map for each unique image. The two forms of GLM employed were an event by event GLM (see \cite{turner2012spatiotemporal}) and a GLM in which individual hemodynamic response functions were estimated for each voxel jointly with the activations \cite{EickenGosa}.
Using this approach we can restrict ourselves to predicting a one-dimensional activity per voxel and image instead of a whole time course.

\subsection{Decoding}
In order to roughly evaluate the information encoded in the preprocessed volumes, we checked whether we could successfully decode the presented texture class using simple linear or generalized linear predictors. Over all subjects the performance of an \(\ell_2\) penalized logistic regression used in a one vs. rest setting cross validated over acquisition blocks was between 70\% and 85\% classification rate with chance level at \(\frac{1}{6}\). This was taken as an indicator that the preprocessed data contained a considerable amount of information about the texture classes.

\subsection{The Scattering Transform}
Using a pyramid of oriented Morlet filters which sufficiently covers the Fourier space of the texture images in a Littlewood-Payley sense, we performed a 2D scattering transform on the images with different parameter settings. Letting \(M\in\{1, 2\}\) represent the number of scattering layers, \(J\in\{4, 5, 6\}\) the number of wavelet scales counting up from the finest and \(L\in\{2, 4, 6, 8\}\) be the number of orientations, we have evaluated all possible combinations: For a given L, let \(\Gamma\) represent the set of \(L\) angles, starting from \(0^\circ\) (horizontal orientation) and moving in steps of \(\frac{180 ^\circ}{L}\). For each \(\gamma\in\Gamma\) let \(\psi_\gamma(x)\) be the Morlet wavelet oriented in \(\gamma\) direction. 
We scale for \(j = 0\) to \(j = J\) as follows: \(\psi_{\gamma, j}(x) = \frac{1}{2 ^ j}\psi_\gamma(2^{-j}x)\). At each layer, a translation invariant scattering network works as follows: For an incoming signal \(u(x)\), the mean \(\int u(x)\textrm{d}x\) is sent to output and a wavelet modulus \((W_{\gamma, j}u)(x) = |\psi_{\gamma, j}\ast u|(x)\) is sent to the next layer. Concretely, for an incoming image \(u_0\), the zeroth output is its mean \(\int u_0(x)\textrm{d}x\). Its wavelet modulus on all scales and orientations \(((W_{\gamma_1, j_1}u_0)(x))_{j_1\leq J, \gamma_1\in\Gamma}\) is propagated to the next layer. At the first layer, we output their means, i.e. \[\int (W_{\gamma_1, j_1}u_0)(x)\textrm{d}x = \int |\psi_{\gamma_1, j_1}\ast u_0|(x)\textrm{d}x\] and propagate their wavelet transforms
\[(W_{\gamma_2, j_2}W_{\gamma_1, j_1}u_0)(x) = |\psi_{\gamma_2, j_2}\ast|\psi_{\gamma_1, j_1}\ast u_0||(x)\]
to level two, where the mean of \((W_{\gamma_2, j_2}W_{\gamma_1, j_1}u_0)(x)\) is sent to output, and so on. It can be shown that the signal energy contained in layer 2 output coefficients having \(j_2 \leq j_1\) is negligible \cite{MallatMathPaper}. Hence only \(j_2 > j_1\) is calculated.

After the scattering transform is done, we thus end up with a mean of the incoming image (layer 0), means of the wavelet transform moduli for all directions \(\gamma_1\in\Gamma\) and scales \(j_1\leq J\) (layer 1) and means of all wavelet transform moduli of wavelet transform moduli for all combinations \(\gamma_2, \gamma_1\in \Gamma\) and \(j_1 < j_2\leq J\) (layer 2). With respect to the number of pixels in the incoming images, this is generally a very small number of coefficients.

\subsection{Encoding model using the Scattering Transform}

We evaluate the predictive power of the translation invariant scattering transform on voxel responses to the presented textures using ridge regression and predictive \(r^2\) scoring. Holding back one of the six acquisition sessions at a time we fit a cross-validated ridge estimator to the rest of the data. The penalty parameter \(\lambda > 0\) of the ridge regression is thus again set using left out sessions: For a given \(\lambda > 0\) we fit the ridge estimator
 on 4 sessions and predict the 5th. The best \(\lambda\) in mean over all 5 folds is chosen to predict the outer left out fold for the 6 outer level scores. The final score is the mean of the 6 outer layer scores. This validation scheme, sometimes called \textit{nested cross-validation}, makes it possible to set parameters without overfitting the data.

\section{Results}

\begin{figure}
  \begin{centering}
    \includegraphics[width=.98\linewidth]{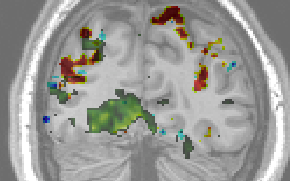}
    \end{centering}
    \begin{centering}
    \includegraphics[width=.98\linewidth]{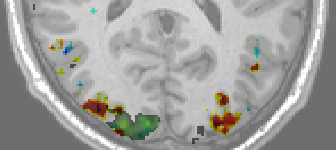}
    \includegraphics[width=.98\linewidth]{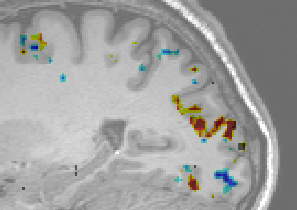}
  \end{centering}
  \caption{Slices showing the encoding performance: Red indicates better modelling by second order features, blue indicates no performance gain using second order features. Green indicates areas that responded to a retinotopy experiment. Top: coronal view. Middle: axial view. Bottom: Sagittal view}\label{fig:m2_m1_diff_texture}
\end{figure}

\begin{figure}
  \begin{centering}
    \includegraphics[width=.9\linewidth]{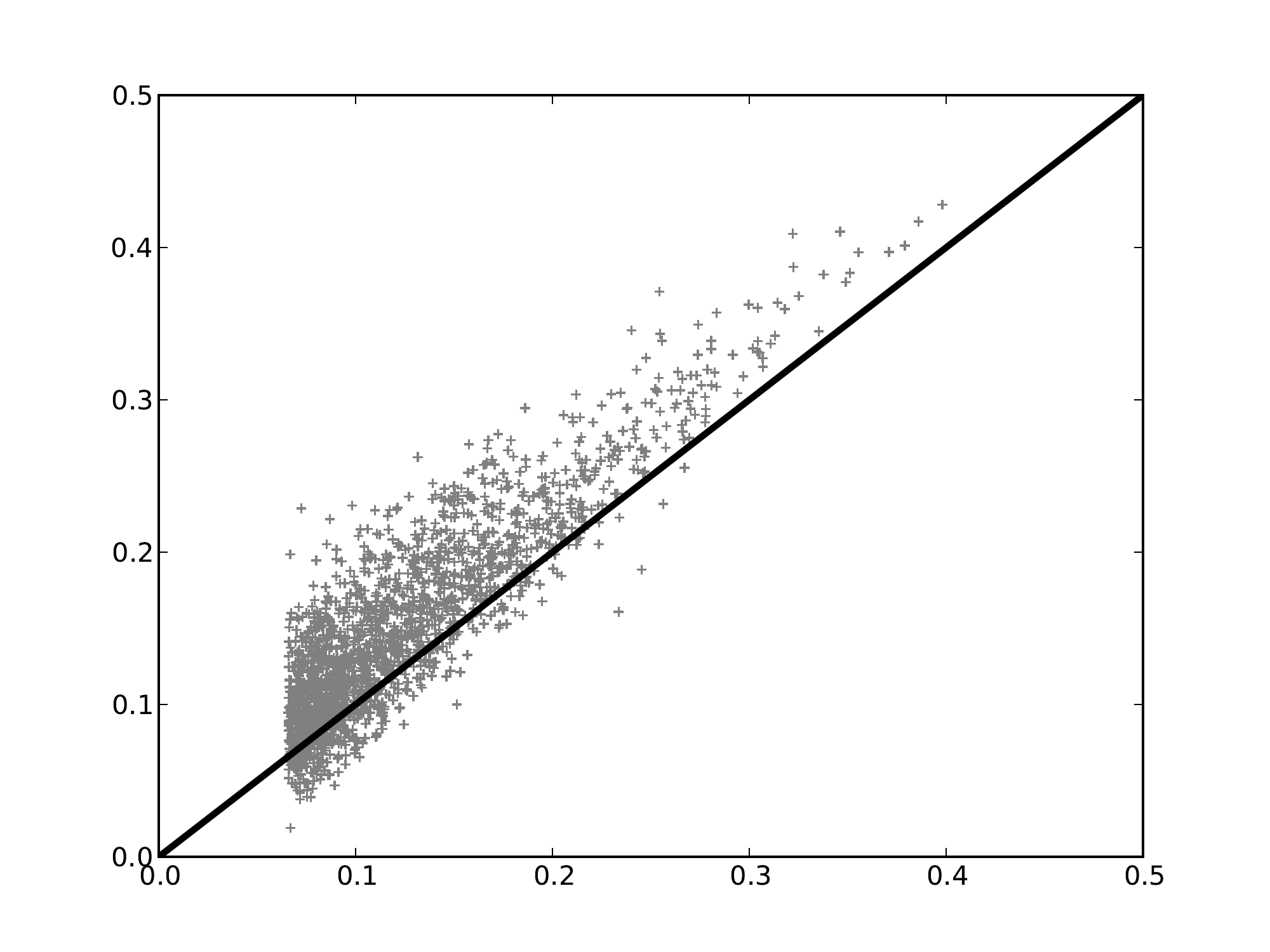}
  \end{centering}
  \caption{Scatterplot showing the 2000 best explained voxels using first order features. These voxels are in  overwhelming majority better explained by second order features.}\label{fig:scatter}
\end{figure}

We find that an overwhelming majority of the voxels that are well-modelled using first order invariant scattering descriptors is largely better modelled using second order invariant scattering descriptors. The scatterplot \ref{fig:scatter} indicates this clearly. A Wilcoxon signed rank test provides a confirmation (\(p < 10^{-5}\)). Figure \ref{fig:m2_m1_diff_texture} shows the localisations of well modelled voxels together with regions that display a high sensitivity to retinotopy. 
One observes that many voxels outside the low-level ventral visual areas are well modelled by the scattering transform. This includes dorsal regions (up to the intra-parietal sulcus), lateral occipital and medio-temporal regions, and not necessarily retinotopic regions; more precisely several foci are found in V1/V2, and more so in V3 and V4. Blue indicates regions where adding second order features to the first order features does not result in a predictive performance gain. Red indicates regions where one observes a performance gain by adding second order features. The regions were coloured red if the performance gain was greater than \(5\%\) of explained variance and blue if the performance gain was \(0\) or performance decreased.
We noticed that the regions that display a strong sensitivity to layer
2, as well as those that actually discriminate between textures, are
confined to some foci, the location of which is highly reproducible
across sessions and even across subjects.

\section{Conclusion}
We have devised an experimental paradigm for studying representations of visual textures in human cortex using fMRI. Using an invariant scattering transform of the stimulus images we built an encoding model and show that voxel activity is fit significantly better if one adds second order descriptors to first order descriptors. This shows that analogously to their beneficial nature to computational texture classification, second order coefficients play a helpful role in fitting voxel activity - independently of the fact whether the modelled brain area is retinotopic.


\section*{Acknowledgment}

The authors would like to thank...
more thanks here



\bibliographystyle{IEEEtran}
\bibliography{vision}

\begin{thebibliography}{10}
\providecommand{\url}[1]{#1}
\csname url@samestyle\endcsname
\providecommand{\newblock}{\relax}
\providecommand{\bibinfo}[2]{#2}
\providecommand{\BIBentrySTDinterwordspacing}{\spaceskip=0pt\relax}
\providecommand{\BIBentryALTinterwordstretchfactor}{4}
\providecommand{\BIBentryALTinterwordspacing}{\spaceskip=\fontdimen2\font plus
\BIBentryALTinterwordstretchfactor\fontdimen3\font minus
  \fontdimen4\font\relax}
\providecommand{\BIBforeignlanguage}[2]{{%
\expandafter\ifx\csname l@#1\endcsname\relax
\typeout{** WARNING: IEEEtran.bst: No hyphenation pattern has been}%
\typeout{** loaded for the language `#1'. Using the pattern for}%
\typeout{** the default language instead.}%
\else
\language=\csname l@#1\endcsname
\fi
#2}}
\providecommand{\BIBdecl}{\relax}
\BIBdecl

\bibitem{HubelWiesel1968}
Hubel and Wiesel, ``Receptive fields and functional architecture of monkey
  striate cortex,'' \emph{Journal of Physiology}, 1968.

\bibitem{EickenbergPRNI2012}
M.~Eickenberg, A.~Gramfort, and B.~Thirion, ``Multilayer scattering image
  analysis fits fmri activity in visual areas,'' in \emph{PRNI2012}, 2012.

\bibitem{EickenbergICMLSTAMLINS2012}
------, ``Scattering transform layer one linearizes functional mri activation
  in visual areas,'' in \emph{ICML STAMLINS Workshop 2012}, 2012.

\bibitem{Kay2008}
K.~N. Kay, T.~Naselaris, R.~J. Prenger, and J.~L. Gallant,
  ``\BIBforeignlanguage{eng}{Identifying natural images from human brain
  activity.}'' \emph{\BIBforeignlanguage{eng}{Nature}}, vol. 452, no. 7185, pp.
  352--355, Mar 2008.

\bibitem{Dalal2005}
Dalal and Triggs, ``Histograms of oriented gradients for human detection,''
  \emph{CVPR IEEE}, 2005.

\bibitem{daisy2007}
E.~Tola, V.~Lepetit, and P.~Fua, ``A fast local descriptor for dense
  matching,'' \emph{Technical Report \'Ecole Polytechnique F\'ed\'erale de
  Lausanne, CVLAB}, 2007.

\bibitem{Portilla2000}
J.~Portilla and E.~P. Simoncelli, ``A parametric texture model based on joint
  statistics of complex wavelet coefficients.'' \emph{International Journal of
  Computer Vision}, pp. 49--70, 2000.

\bibitem{Bruna2011}
J.~Bruna and S.~Mallat, ``Classification with scattering operators,'' in
  \emph{CVPR}.\hskip 1em plus 0.5em minus 0.4em\relax IEEE, 2011, pp.
  1561--1566.

\bibitem{Bruna2012}
------, ``Invariant scattering convolution networks,'' in \emph{PAMI}, 2012.

\bibitem{Sifre2012}
L.~Sifre and S.~Mallat, ``Combined scattering for rotation invariant texture
  analysis,'' in \emph{ESANN}, 2012.

\bibitem{KFlats}
\BIBentryALTinterwordspacing
G.~Canas, D., T.~Poggio, and L.~Rosasco, ``Learning manifolds with k-means and
  k-flats,'' 2012, to appear in Advances in Neural Information Processing
  Systems, NIPS 2012. [Online]. Available: \url{http://arxiv.org/abs/1209.1121}
\BIBentrySTDinterwordspacing

\bibitem{Naselaris2011}
\BIBentryALTinterwordspacing
T.~Naselaris, K.~N. Kay, S.~Nishimoto, and J.~L. Gallant,
  ``\BIBforeignlanguage{eng}{Encoding and decoding in fmri.}''
  \emph{\BIBforeignlanguage{eng}{Neuroimage}}, vol.~56, no.~2, pp. 400--410,
  May 2011. [Online]. Available:
  \url{http://dx.doi.org/10.1016/j.neuroimage.2010.07.073}
\BIBentrySTDinterwordspacing

\bibitem{Lazebnik05asparse}
S.~Lazebnik, C.~Schmid, and J.~Ponce, ``A sparse texture representation using
  local affine regions,'' \emph{IEEE Transactions on Pattern Analysis and
  Machine Intelligence}, vol.~27, pp. 1265--1278, 2005.

\bibitem{Boegle2010}
\BIBentryALTinterwordspacing
R.~Boegle, J.~Maclaren, and M.~Zaitsev,
  ``\BIBforeignlanguage{English}{Combining prospective motion correction and
  distortion correction for epi: towards a comprehensive correction of motion
  and susceptibility-induced artifacts},''
  \emph{\BIBforeignlanguage{English}{Magnetic Resonance Materials in Physics,
  Biology and Medicine}}, vol.~23, no.~4, pp. 263--273, 2010. [Online].
  Available: \url{http://dx.doi.org/10.1007/s10334-010-0225-8}
\BIBentrySTDinterwordspacing

\bibitem{turner2012spatiotemporal}
B.~O. Turner, J.~A. Mumford, R.~A. Poldrack, and F.~G. Ashby, ``Spatiotemporal
  activity estimation for multivoxel pattern analysis with rapid event-related
  designs,'' \emph{NeuroImage}, 2012.

\bibitem{EickenGosa}
F.~Pedregosa, M.~Eickenberg, B.~Thirion, and A.~Gramfort, ``Hrf estimation
  improves sensitivity of fmri encoding and decoding models,'' in
  \emph{PRNI2013, submitted}, 2013.

\bibitem{MallatMathPaper}
S.~Mallat, ``Group invariant scattering,'' \emph{to appear in Comm. Pure and
  Appl. Math.}, 2011.

\end{thebibliography}
%
%

\end{document}